\documentclass[letterpaper, 10pt, conference]{ieeeconf}      

\IEEEoverridecommandlockouts                              

\usepackage{graphics} 
\usepackage{epsfig} 
\usepackage{mathptmx} 
\usepackage{times} 
\usepackage{amsmath} 
\usepackage{amssymb}  
\usepackage[colorlinks,
            linkcolor=blue,       
            anchorcolor=blue,  
            citecolor=blue,        
            ]{hyperref}
\usepackage{graphicx}
\usepackage{stfloats}
\usepackage{multirow}

\title{\LARGE \bf
AdvSwap: Covert Adversarial Perturbation with High Frequency Info-swapping for Autonomous Driving Perception
}

\author{Yuanhao Huang$^{1}$, Qinfan Zhang$^{1}$,Jiandong Xing$^{1}$, Mengyue Cheng$^{1}$, Haiyang Yu$^{2}$, Yilong Ren$^{2,*}$\\ and Xiao Xiong$^{3}$
\thanks{$^{1}$Yuanhao Huang, Qinfan Zhang, Jiandong Xing, and Mengyue Cheng are with the School of Transportation Science and Engineering, Beihang University, Beijing, 100191, P.R.China, and State Key Lab of Intelligent Transportation System, Beijing, 100191,P.R.China.
        {\tt\small yuanhao\_huang@buaa.edu.cn}}%
\thanks{$^{2}$Haiyang Yu and Yilong Ren are with the School of Transportation Science and Engineering, Beihang University, Beijing, 100191, China, and Zhongguancun Laboratory, Beijing 100094, P.R.China. The corresponding author is Yilong Ren.
        {\tt\small yilongren@buaa.edu.cn}}%
\thanks{$^{3}$Xiao Xiong are with the Department of Electrical and Computer Engineering,University of Alberta, Edmonton, Canada
        {\tt\small xiongxiao917@gmail.com}}%
}

\begin{document}

\maketitle
\thispagestyle{empty}
\pagestyle{empty}

\begin{abstract}

Perception module of Autonomous vehicles (AVs) are increasingly susceptible to be attacked, which exploit vulnerabilities in neural networks through adversarial inputs, thereby compromising the AI safety. Some researches focus on creating covert adversarial samples, but existing global noise techniques are detectable and difficult to deceive the human visual system. This paper introduces a novel adversarial attack method, AdvSwap, which creatively utilizes wavelet-based high-frequency information swapping to generate covert adversarial samples and fool the camera. AdvSwap employs invertible neural network for selective high-frequency information swapping, preserving both forward propagation and data integrity. The scheme effectively removes the original label data and incorporates the guidance image data, producing concealed and robust adversarial samples. Experimental evaluations and comparisons on the GTSRB and nuScenes datasets demonstrate that AdvSwap can make concealed attacks on common traffic targets. The generates adversarial samples are also difficult to perceive by humans and algorithms. Meanwhile, the method has strong attacking robustness and attacking transferability.


\end{abstract}

\section{INTRODUCTION}

Autonomous driving, a cutting-edge technology ensures the safety and efficient transportation~\cite{bai2024ar, zhang2022intelligent}, is progressing rapidly. However, similar to lots of intelligent system, it is susceptible to malicious attacks\cite{cai2024adversarial}. Adversarial attacks on autonomous vehicles, specifically targeting the visual perception systems, have been a growing concern in recent researches~\cite{kong2020physgan}. The attacks aim to deceive the vehicles by manipulating the inputs from cameras, potentially leading to dangerous consequences\cite{huang2022practical}. Attacks on the images can be broadly categorized into two types: physical attacks and digital attacks, with the exact distinction depending on their order relative to when the image was captured.

Physical attacks involve the placement of specially crafted attack patches in the physical environment before imaging. The patches are designed to deceive the perception algorithms of autonomous vehicles, leading to compromised accuracy and potentially hazardous situations such as false detection or missed detection. For example, ~\cite{ye2021patch} proposed an adversarial patch-based method to attack the recognition of traffic signs. On the other hand, digital attacks occur after imaging of camera. The digital attacks target the pixel information within the image and are characterized by their ability to conceal the manipulation while maintaining high efficiency in fooling the perception algorithms~\cite{wei2021black}.

\begin{figure}[t]
\centering
\includegraphics[width=7.5 cm]{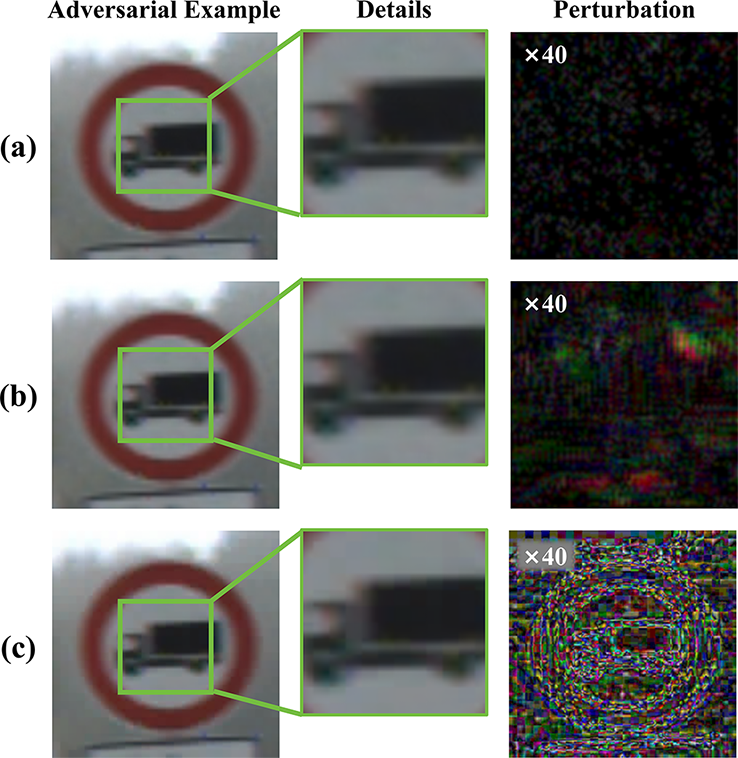}
\caption{Perturbation comparison between the generated image and the original image. Adversarial examples generated by three algorithms: (a) Proposed AdvSwap, (b) SSAH~\cite{luo2022frequency}, (c) AdvDrop~\cite{duan2021advdrop}.}
\label{1_img_show}
\vspace{-6 mm} 
\end{figure}

In the field of AVs, digital attacks targeting perceived images aim to disrupt or deceive the perception algorithms of autonomous driving systems. The attacks pose a serious threats as they can lead to incorrect the decisions and behaviors of AVs covertly~\cite{ye2021patch}. For example, methods for generating adversarial samples introduce subtle changes, such as gradient noise, to the original image to deceive the perception algorithms. ~\cite{yan2023adversarial} developed a robust adversarial perturbation method targeting salient image regions with class activation mapping, which enable confounding effects maximization via front-door intermediates. ~\cite{chow2020adversarial} presented a suite of adversarial objects gradient attacks, coined as TOG, which can cause the state-of-the-art recognition algorithms to suffer from untargeted random perturbation. In summary, transformations to the original image include adding noise, modifying pixel values and making small perturbation. 

Nevertheless, perturbations generated by Adversarial method might manifest in regions noticeable to the human visual system (HVS)~\cite{lopez2021survey}. Another creative method involves adding or dropping adversarial targets from the image or directly tampering with image pixels to get low perceptual losses. The methods can directly influence the perception algorithm's decision-making process, thereby increasing its potential deception and effectiveness. As a novel adversarial attack, AdvDrop crafts adversarial examples by dropping imperceptible details' information from images~\cite{duan2021advdrop}. 

Meanwhile, natural noise is typically a random and irregular perturbation that primarily affects low-frequency features of an image. Researchers have focused on  the algorithms that targets specific pixels in the region of high-frequency~\cite{chen2023imperceptible}. SSAH introduce the low frequency constrain to limit perturbations within high frequency components, ensuring perceptual similarity between adversarial examples and originals~\cite{luo2022frequency}. 

In this paper, we propose a covert adversarial attack method for autonomous driving images, called AdvSwap. The method achieves efficient and covert adversarial attacks by exchanging high-frequency information in the frequency domain and designing an optimizer. In the process of high-frequency information swapping, the invertible network module is relied on forward propagation to ensure that the information of the data before and after wavelet transformation is completely retained. The same amount of information exchange can delete the semantic information of the original image label and add the information of the target label to form an adversarial sample. The main contributions can be summarized as follows:

\begin{itemize}

\item [1)] Proposed a novel adversarial method with information swapping module based on wavelet transform and invertible network module. The method aim to generate covert adversarial sample by adding and deleting same amount of information.

\item [2)]	A adversarial optimizer is designed to learn the noise information of the target image as a guide; a classification optimizer is also designed to guide high-frequency information exchange to improve the attack success rate with less disturbance.

\item [3)]	We conducted lots of experiments on the GTSRB and nuScenes datasets and compared them with two advanced algorithms. We proved that this method can efficiently, robustly and covertly attack traffic object images.

\end{itemize}


\section{Related Works}

\subsection{Adversarial Attacks}

\subsubsection{Digital Attacks}

In the study of adversarial attacks, Digital Attacks are the most basic and widely used method. They mainly add small but intentional perturbations to the input data of machine learning models, especially deep neural networks. Early representative work such as the Fast Gradient Sign Method (FGSM) proposed by Goodfellow et al~\cite{goodfellow2014explaining}. By adding small-amplitude perturbations along the gradient direction to the original samples, it misleads the model to produce wrong predictions. Subsequent research has developed a series of more sophisticated and powerful attack methods, such as AdaBelief FGSM (AB-FGSM)~\cite{wang2022ab}, Projected Gradient Descent (PGD)~\cite{gupta2018cnn}, etc., which further optimize the perturbation. The addition method improves the concealment and attack success rate of adversarial samples.

\subsubsection{Adding and Droping of Perturbation}

The method of adding and removing perturbations is a typical strategy for generating adversarial examples. The DeepFool~\cite{moosavi2016deepfool} attack generates adversarial examples by minimizing the amount of perturbation that needs to be added from the original sample to the decision boundary. In contrast, the Universal Perturbation method~\cite{moosavi2017universal} focuses on finding a universal perturbation vector that can produce adversarial effects on a large number of samples. In addition, the generation of adversarial samples can also be achieved by removing certain key information. For example, AdvDrop~\cite{duan2021advdrop} generates high-quality adversarial samples by discarding high-frequency parts of the image. Together, these methods reveal the flexibility of adversarial attacks in adding and removing perturbations and diversity.

\subsubsection{Covert Attacks}

Covert Attacks focus on generating adversarial samples that are difficult to detect or even imperceptible to bypass existing defense algorithms. For example, Semantic Attacks~\cite{joshi2019semantic} focus on generating adversarial examples while keeping the semantics of the samples unchanged, making it difficult for humans to detect the changes. Additionally, SSAH~\cite{luo2022frequency} is a novel algorithm that attacks semantic similarities on feature representations.

\subsection{Information Swapping}

\subsubsection{Wavelet Decomposition}

Wavelet decomposition is a traditional image processing method that can decompose images into feature details at different scales and directions in the frequency domain. By performing adversarial perturbation operations in the wavelet domain, the attacker is able to embed covert adversarial information while preserving the overall structure of the original image~\cite{yan2023wavelet}. Generating the adversarial examples in this way can effectively mislead the perception model while the examples are visually close to the original examples~\cite{luo2022frequency}.

\subsubsection{Residual Dense Network}

Residual Dense helps the network learn the difference between the input and the expected output when constructing adversarial samples~\cite{zhang2018residual}. The goal of network design is to efficiently construct adversarial samples while retaining some of the discriminative features of the source image and injecting semantic features unique to the target category.

\subsubsection{Invertible Block}

Invertible neural networks (INNs) have garnered significant attention due to their ability to construct reversible modules for steganographic purposes~\cite{chen2023imperceptible}. With INNs, researchers can hide host image information whthin container image files, and even hide information from adversarial samples to mislead detection algorithms. For instance, ~\cite{lu2021large} apply invertible block to adversarial attacks, the perturbations can deceive deep learning models, leading to misclassification or erroneous outputs.


\begin{figure*}[htbp!]
\centering
\includegraphics[width=16 cm]{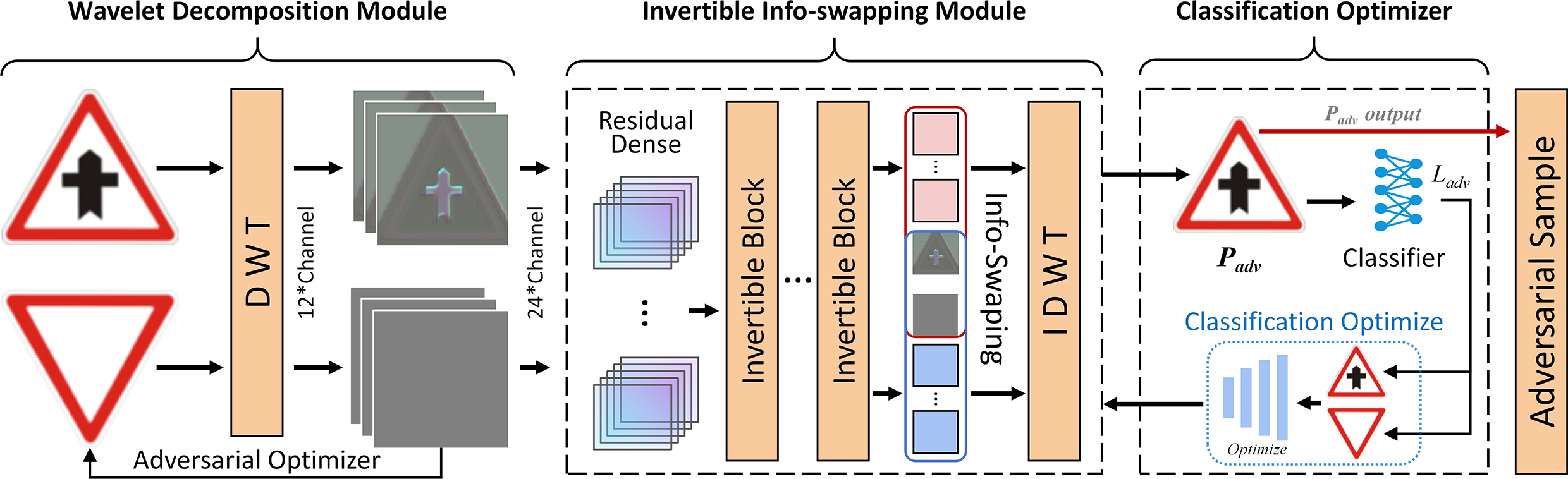}
\caption{Overview of the proposed adversarial attack method.}
\label{2_structure}
\vspace{-6 mm} 
\end{figure*}

\section{Proposed Method}

\subsection{Overview}

Given an original clear image $x_{orig} \in \mathbb{R}^n$, which is labeled $y$, the goal of the attack is to convert it into another image $x_{\text{adv}}$ by the adversarial example generation algorithm, whose label is $c \neq y$, where $c$ is the wrong label that the attacker hopes to mislead the classifier to output. The adversarial attack algorithm framework proposed in this article involves a pre-trained classifier $f: \mathbb{R}^n \rightarrow \{1,2,...,n\}$, which can convert the input The image pixel information is classified into a specific category among $n$ categories.

The process of adversarial sample generation is regarded as an optimization problem, and hidden adversarial samples $x_{\text{adv}}$ are generated by solving a specific optimization loss function. Specifically, the eq.\ref{eq1} defines such an optimization goal, which is to find the parameter $\theta$ that minimizes the loss while satisfying the constraints:

\begin{equation}
x_{\text{adv}} = \arg\underset{\theta}{\min} \, \left[ \lambda_{\text{adv}} \mathcal{L} _{\text{adv}}(x_{\text{adv}}, c) + \mathcal{L} _{\text{swap}}(x_{\text{adv}}, x_{\text{orig}}) \right] 
\label{eq1}
\end{equation}

Here, $\lambda_{\text{adv}}$ is the regularization parameter of the adversarial loss $\mathcal{L} _{\text{adv}}$, which measures the adversarial sample $x_{\text{adv} }$ is the distance between the expected misleading category $c$. $\mathcal{L} _{\text{swap}}$ is the reconstruction loss, which ensures that the generated adversarial sample retains the visual characteristics of original image $x_{text{orig}}$.

And it also needs to satisfy the following constraint to ensure imperceptibility:

\begin{equation}
s.t. \ \|x_{\text{adv}} - x_{\text{orig}}\|_{\infty} \leq \epsilon
\label{eq2}
\end{equation}

This means that the maximum difference (infinite norm $l_\infty$) between the generated adversarial sample $x_{\text{adv}}$ and the original clean sample $x_{\text{cln}}$ does not exceed a small constant $\epsilon $. The overview of AdvSwap is illustrated in Fig. \ref{2_structure}. The structure consists of Wavelet Decomposition Module, Invertible Info-swapping Module, and Classification Optimizer.

\subsection{Wavelet Decomposition Module}

Unlike pixel-space adversarial attacks, manipulations conducted in the frequency domain with discrete wavelet transform (DWT), frequently yield less perceptible perturbations. When subjected to DWT, an input RGB image  $x \in \mathbb{R}^{C \times H \times W}$ , where  $C=3$ , is transformed into its wavelet domain representation  $T(x) \in \mathbb{R}^{4C \times \frac{H}{2} \times \frac{W}{2}}$ , featuring one low-frequency sub-band and three high-frequency sub-bands for each color channel. The 12 distinct frequency sub-bands include the approximation coefficients  $\phi$ , horizontal details  $\varphi$ , vertical details  $\rho$ , and diagonal details  $\eta$ , capturing different spatial-frequency properties. 

The wavelet-based representation facilitates precise manipulation for creating stealthy adversarial examples, compatible with subsequent processing or analysis. Following the necessary manipulations, the Inverse Discrete Wavelet Transform (IDWT), denoted as  $T^{-1}(\cdot)$ , is employed to reconstruct the features back into the image domain, maintaining the stealthiness of the adversarial attack.

\subsection{Invertible Info-swapping Module}

\begin{figure}[htbp!]
\centering
\includegraphics[width=6.5 cm]{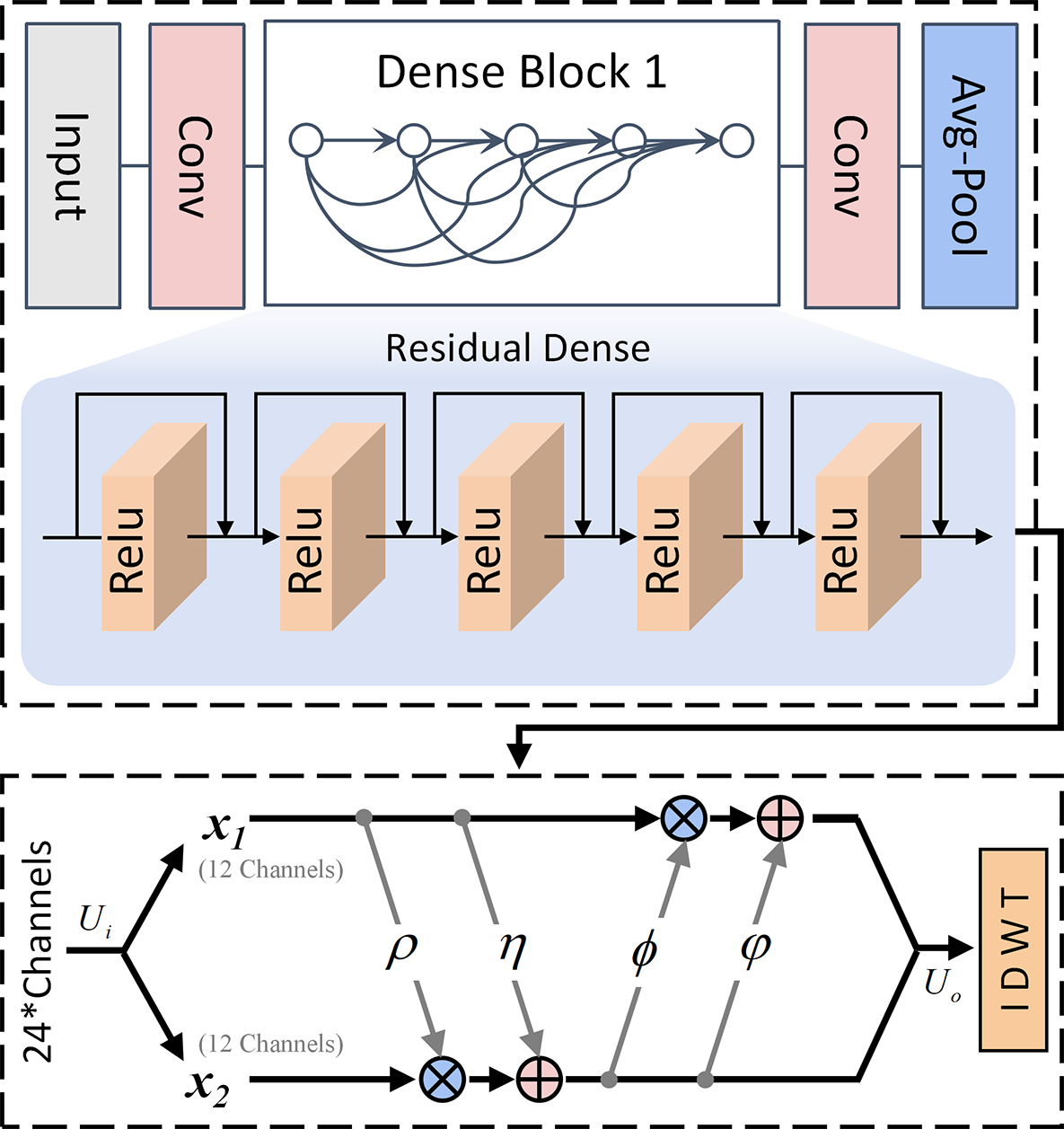}
\caption{The reversible module exchanges wavelet component information for feature extraction and enables deletion of original image information while injecting covert adversarial perturbation.}
\label{3_info_swapping}
\end{figure}

As shown in Fig.\ref{3_info_swapping}, Invertible Info-swapping Module is build by multi-residual networks, invertible blocks, and info-swapping blocks. DWT decomposes the image into different component characteristics, ensuring that the adversarial algorithm can carry out covert information tampering in high-frequency information. The reversibility of the Invertible block ensures the integrity of the original data during adversarial information hiding. 

\subsubsection{\textbf{Invertible Info-swapping Module}}

Building a reversible module based on the invertible neural network\cite{ardizzone2021conditional} can achieve complete retention and recovery of information during the data conversion process. Assume that $ w_{i-1_{cln}} $ is the cleaning feature input in the $ (i-1) $th Invertible Block, and $ w_{i-1_{tgt}} $ is the same block input target features. Each Invertible Block alternates which part of the input it conditions the scaling and translation on, allowing for information to flow between the two sets of features ($w_{cln}$ for clean data and $w_{tgt}$ for target data).:

\begin{equation}
\boldsymbol{\tilde{w}}_{c l n}^{(i)} = \boldsymbol{w}_{c l n}^{(i-1)} \otimes \exp \left( \beta \left( \mathbb{F} \left(\boldsymbol{w}_{t g t}^{(i-1)}\right) \right) \right) +  \mathbb{F} \left(\boldsymbol{w}_{t g t}^{(i-1)}\right),
\label{eq3}
\end{equation}

where $\otimes$ denotes element-wise multiplication, and $\beta$ represent a Sigmoid function multiplied by a constant factor. $\mathbb{F} $ denotes the residual dense network for each wavelet sub-bands as shown in \cite{zhang2018residual}.

\subsubsection{\textbf{Information Swapping}}

The inverse operation of process is essential for invertible neural networks, where the transformation must be reversible. For the $i$-th Info-swapping, the inverse operation would be:

\begin{equation}
\begin{split}
w^{i-1}_{cln} &= (w^i_{cln} - t^i_{cln}) \otimes exp(-s^i_{cln}), \\
w^{i-1}_{tgt} &= (w^i_{tgt} - t^i_{tgt}) \otimes exp(s^i_{tgt}).
\label{eq4}
\end{split}
\end{equation}

The module allow for efficient bi-directional transformations and enable exact likelihood computation in generative models.

The defined inverse operations in Eq. (\ref{eq4}) guarantee the reversibility of the info-swapping process, which is fundamental for retaining objective mappings and precise likelihood estimation in generative models. To further enhance the creation of imperceptible adversarial instances while maintaining the visual integrity post-swapping, we introduce a swapping loss function $ \mathcal{L}_{swap} $:

\begin{equation}
\begin{split}
\mathcal{L}_{swap} =\sum_{i \in \{\rho, \eta, \phi, \varphi\}} &\lambda_i \lVert P(x_{orig})_i - P(x_{adv})_i \rVert_2^2 + \\
&\lambda_{perp} \lVert \rho(x_{orig})_i - \rho(x_{adv})_i \rVert_2^2,
\label{eq5}
\end{split}
\end{equation}

where $ P(\cdot) $ denotes the operator extracting the discrete wavelet transform coefficients across sub-bands LL, LH, HL, and HH. $ \rho(\cdot) $ refers to an auxiliary feature extractor capturing perceptually significant attributes. $ \lambda_i $ are scalar coefficients weighing the contribution of each wavelet coefficient to the reconstruction fidelity. $ \lambda_{perp} $ is a weight parameter accounting for the loss in high-level feature consistency between the original and adversarial instances.

This compact loss formulation consolidates discrepancies in wavelet decomposition between the original and adversarial images and integrates a penalty term for preserving high-level perceptual congruence.

\subsection{Adversarial and Classification Optimizer}

\subsubsection{\textbf{Adversarial Optimizer}}

In this segment, we discuss the Adversarial Optimizer, tasked with creating a carefully manipulated noise pattern, the Target Guide Sample (TGS). Initially set to a uniform half-intensity image ($ I_{tgs}(i,j,k) = 0.5 $, where $ I_{tgs} $ is of dimensions $ H \times W \times C $), the TGS experiences stochastic iterative perturbations to its pixels.

The optimizer leverages a gradient-based mechanism, updating pixels according to Eq.~\ref{eq6}:

\begin{equation}
I_{tgs}'(i,j,k) = I_{tgs}(i,j,k) + \epsilon \cdot sign(\nabla_{I_{tgs}(i,j,k)} \mathcal{L} (f(I_{tgs}), y_t)),
\label{eq6}
\end{equation}

where $ I_{tgs}' $ signifies the perturbed TGS, $ f $ is the attacked classification model, $ y_t $ is the target class label, $ \mathcal{L}  $ is the loss function—commonly cross-entropy—and $ \epsilon $ controls perturbation strength. The sign function guides the gradient updates towards maximizing the classifier's misclassification confidence.

Crucially, $ \mathcal{L} $ measures the mismatch between the classifier's output and the target. As the classifier's confidence in assigning the target class to $ I_{tgs}' $ increases, indicated by $ p(y_t | I_{tgs}') $, the loss $ \mathcal{L} (f(I_{itgs}'), y_t) $ diminishes. Once the confidence exceeds a preset threshold, denoting successful adversarial manipulation, the optimization halts. This method showcases the Adversarial Optimizer's proficiency in crafting imperceptible yet deceptive perturbations, compelling the classifier to classify the noise as the target class with high certainty.

\subsubsection{\textbf{Classification Optimizer}}

The Classification Optimizer module gauges the result of the $ P_{adv} $ adversarial sample. It measures the discrepancy in classification outcomes between adversarial images $ P_{adv} $ and their respective Target Guide Sample (TGS) labels using Cross-Entropy Loss:

\begin{equation}
\mathcal{L}_{adv}(P_{adv}, y_t) = -y_t \cdot \log(p_{model}(y_t | P_{adv}))
\label{eq6}
\end{equation}

where $ y_t $ is the target class label's one-hot representation, and $ p_{model}(y_t | P_{adv}) $ is the classifier's output probability for that target class given the adversarial input. This loss metric quantifies how closely the classifier adheres to the adversary's intended misclassification; minimization implies greater performers in generating an effective adversarial example.


\section{Experiments and Discussion}

\begin{figure*}[hb!]
\centering
\includegraphics[width= 16.7 cm]{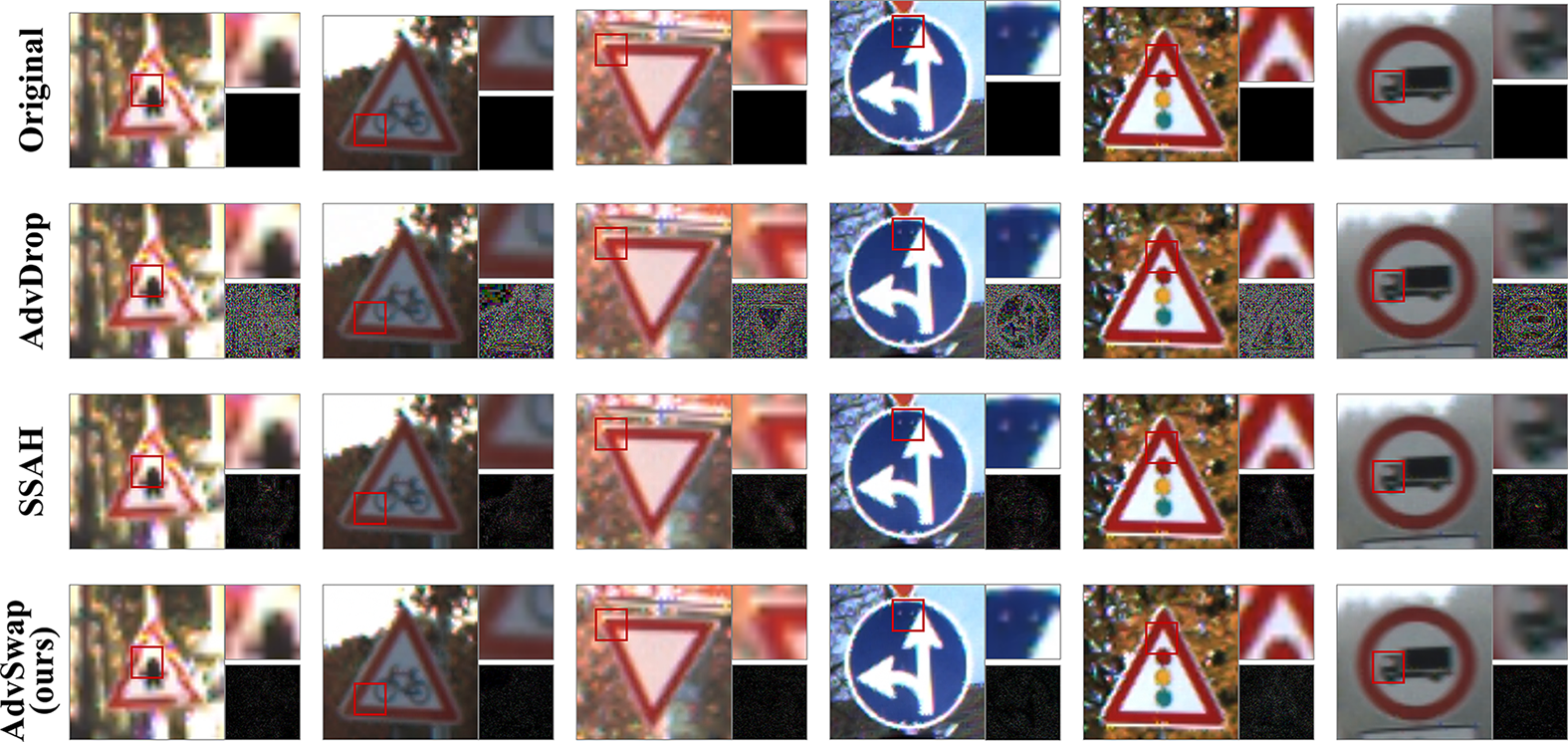}
\caption{Comparison of adversarial samples and perturbed images generated by the SSAH, AdvDrop and the proposed AdvSwap in GTSRB.}
\label{4_comparison_GTSRB}
\end{figure*}

\subsection{Experimental Setup}

\subsubsection{\textbf{Testing Datasets}}

We selected 10 images in each of 43 categories from the German Traffic Sign Recognition Benchmark dataset (GTSRB)~\cite{HoubenIJCNN2013}, totaling 430 images. And a total of 1072 pictures in 22 categories were selected in the nuScenes dataset~\cite{nuscenes}. All images can be correctly recognized and classified by the retrained Resnet-18. All methods were verified on these two datasets and trained on a computer equipped with an RTX 3080 GPU.

Set the optimizer of the optimization task in Eq.~\ref{eq1} to Adam. The initial learning rate is $1e^-4$, the decay rate is 0.9 every 200 iterations, and the lower bound is $1e^-5$. At the same time, set other parameters of the algorithm based on experience: $\lambda_{adv}=3$, $\lambda_{\rho}=2$, $\lambda_{\eta, \phi, \varphi}=1$, $\lambda_{perp}=1$, and $\epsilon=8/255$.

\subsubsection{\textbf{Comparison Method}}

We compare proposed method with two recent state-of-the-art methods. One is a novel adversarial attack SSAH~\cite{luo2022frequency} proposed by Cheng Luo et al., which is applicable in wide settings by attacking the semantic similarity of images. Another is AdvDrop~\cite{duan2021advdrop}, which crafts adversarial images by dropping existing details of clean images.

\subsubsection{\textbf{Evaluation Indicators}}

We use multiple quantitative metrics to comprehensively evaluate the effectiveness of algorithm-generated adversarial examples, as follows:

\begin{itemize}

\item Attacking success rate (ARS): The probability of successful detection/missing detection by the detector against an adversarial sample.

\item Mean Square Error (MSE): The average of the squared differences of each pixel value between the processed image and the original image, measuring the pixel-level difference between the two.

\item Peak Signal to Noise Ratio (PSNR)~\cite{hore2010image}: Represents the relative error between the original image and the processed image. The higher the PSNR value, the closer the processed image quality is to the original image.

\item Structural Similarity Index (SSIM)~\cite{wang2004image}: It is used to measure the structural similarity between the original image and the processed image. The closer to 1, the more similar the two images are.

\item Learned Perceptual Image Patch Similarity (LPIPS)~\cite{zhang2018unreasonable}: Using deep learning models to learn image perceptual features to measure the perceptual differences between original images and adversarial examples.

\item $\mathbf{l_2}$-normal: Measures the average Euclidean distance of pixel values between the adversarial example and the original image.

\item $\mathbf{l_\infty}$-normal: Measures the maximum difference in pixel values between the adversarial example and the original image.

\end{itemize}

\subsection{Attacks Experiments}

\begin{figure*}[ht!]
\centering
\includegraphics[width= 16.7 cm]{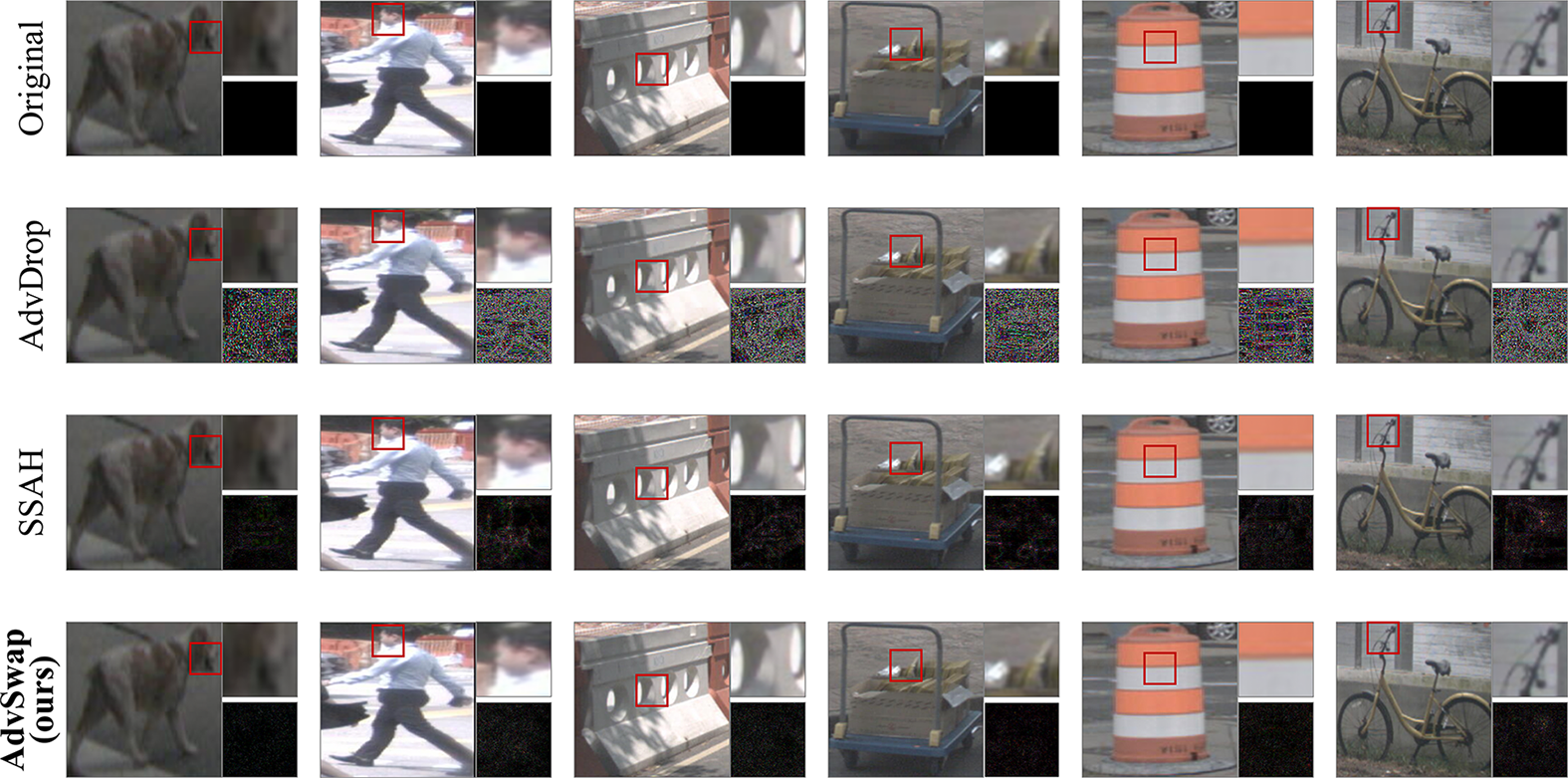}
\caption{Comparison of adversarial samples and perturbed images generated by the SSAH, AdvDrop and the proposed AdvSwap in nuScenes.}
\label{5_comparison_nuScenes}
\vspace{-6 mm} 
\end{figure*}

\begin{table}[ht!]\small
\centering
\renewcommand\arraystretch{1.2}
\caption{ASR and Evaluation Indicators Comparison in GTSRB}
\setlength{\tabcolsep}{2.5mm}{
\begin{tabular}{cccc}
\hline
\multicolumn{4}{c}{\textbf{GTSRB}}                \\ \hline
                & \textbf{SSAH}~\cite{luo2022frequency}   & \textbf{AdvDrop}~\cite{duan2021advdrop} & \textbf{AdvSwap} (ours) \\ \hline
ASR ($\uparrow$)      & 0.935  & 0.956   & \textbf{0.990}  \\
MSE ($\downarrow$)    & \underline{0.007}  & 0.027   & \textbf{0.006} \\
PSNR ($\downarrow$)   & 52.785 & \textbf{41.296}  & 54.575 \\
SSIM ($\uparrow$)     & \underline{0.998}  & 0.975   & \textbf{0.997} \\
LPIPS ($\downarrow$)  & 0.031  & 0.211   & \textbf{0.041}  \\
$l_2$ ($\downarrow$)  & \underline{0.822}  & 12.117  & \textbf{0.585}  \\
$l_\infty$ ($\downarrow$)& \underline{0.020}  & 0.062   & \textbf{0.012} \\ \hline
\end{tabular}}
\label{tab1}
\vspace{-4 mm}
\end{table}

Tab.~\ref{tab1} presents a quantitative comparison of our proposed AdvSwap method against existing techniques SSAH~\cite{luo2022frequency} and AdvDrop~\cite{duan2021advdrop} on the GTSRB dataset. The table reveals that AdvSwap demonstrates strong performance across several key metrics.

AdvSwap achieves a high Attack Success Rate (ASR) of $0.990$, higher than the 2th-performing AdvDrop ($0.956$). Regarding imperceptibility, AdvSwap sets new lows for Mean Squared Error (MSE) at $0.006$ and shows superior Structural Similarity Index Measure (SSIM) of $0.997$, suggesting it introduces minimal visible changes while preserving image structure closer to the original compared to others.

In terms of perceptual similarity, AdvSwap records the least Learned Perceptual Image Patch Similarity (LPIPS) score of $0.041$, indicating a lower perceived distortion. Moreover, AdvSwap outperforms competitors in both $l_2$ and $l_{\infty}$ norms with smaller perturbation sizes: $l_2$ at $0.585$ and $l_{\infty}$ at $0.012$.

Visual qualitative display in Fig.~\ref{4_comparison_GTSRB} provide further evidence, showcasing Original Pictures, Adversarial Pictures, Detail Comparisons, and Perturbation Maps. We can see that AdvSwap is able to generate more visually invisible adversarial instances than the compared algorithms. The possible reason is that the proposed algorithm only interacts with high-frequency information to hide a small amount of semantic information. However, the contrast algorithm uniformly increases the perturbation globally in the picture by limiting the number of perturbations of $l_2$ and $l_\infty$~\cite{luo2022frequency}.

\begin{table}[h!]\small
\centering
\renewcommand\arraystretch{1.2}
\caption{ASR and Evaluation Indicators Comparison in nuScenes}
\setlength{\tabcolsep}{3mm}{
\begin{tabular}{cccc}
\hline
\multicolumn{4}{c}{\textbf{nuScenes}}                \\ \hline
                & \textbf{SSAH}~\cite{luo2022frequency}   & \textbf{AdvDrop}~\cite{duan2021advdrop} & \textbf{AdvSwap} (ours) \\ \hline
ASR ($\uparrow$)      & 0.914  & \textbf{0.991}   & 0.989          \\
MSE ($\downarrow$)    & \underline{0.007}  & 0.027   & \textbf{0.005}          \\
PSNR ($\downarrow$)   & 52.077 & \textbf{39.806}  & 54.229         \\
SSIM ($\uparrow$)     & \underline{0.997}  & 0.966   & \textbf{0.998}          \\
LPIPS ($\downarrow$)  & \underline{0.014}  & 0.168   & \textbf{0.012}          \\
$l_2$ ($\downarrow$)  & \underline{0.986}  & 16.875  & \textbf{0.619}     \\
$l_\infty$ ($\downarrow$) & \underline{0.023}  & 0.079   & \textbf{0.012}      \\ \hline
\end{tabular}}
\label{tab2}
\end{table}

Expanding our experimentation to the nuScenes dataset, which encompasses a diverse range of traffic scenarios with 22 object categories, provides further validation of AdvSwap. This dataset serves as a comprehensive benchmark for object detection systems in autonomous driving applications. Impressively, AdvSwap achieves an ASR of 98.9\%, nearly matching the performance of the top performer, demonstrating its adaptability to complex real-world conditions. Notably, AdvSwap excels in minimizing perturbation visibility, as evidenced by its lowest MSE among competitors and superior SSIM score, highlighting its ability to generate adversarial samples that closely resemble the original scenes. Supplementary visual analyses in associated figures (Fig.~\ref{5_comparison_nuScenes}) provide qualitative evidence. The visualizations underscore the subtle yet potent nature of the perturbations introduced by AdvSwap, which effectively fool object detectors without obvious visual artifacts.

In summary, across the GTSRB and nuScenes datasets, our proposed AdvSwap algorithm demonstrates exceptional performance, achieving high attack success rates while preserving image quality and introducing minimally perceptible perturbations. Outperforming or closely matching state-of-the-art methods in metrics like MSE, PSNR, SSIM, and LPIPS, AdvSwap confirms its versatility and robustness in crafting effective adversarial examples for diverse computer vision tasks, reinforcing its value in enhancing model resilience assessments.

\subsection{Robustness Experiments}

We conducted 2 defense scheme tests (JPEG and Shield~\cite{das2018shield}) against 3 attack methods, and evaluated the robustness of the algorithm through the attack success rate. The assessment was carried out using two benchmark datasets, GTSRB and nuScenes, with the primary performance metric being the recognition accuracy (RA).

\begin{table}[htbp!]\small
\centering
\renewcommand\arraystretch{1.2}
\caption{Robustness experiments under different defense methods}
\setlength{\tabcolsep}{1.5mm}{
\begin{tabular}{cccc}
\hline
\multirow{2}{*}{Defense}                     & \multirow{2}{*}{Adversarial Attack} & \multicolumn{2}{c}{Recognition Accuracy} \\ \cline{3-4} 
                                             &                                     & GTSRB                 & nuScenes              \\ \hline
\multirow{3}{*}{No Defend}                   & AdvDrop                             & 6.49\%                & 3.69\%                \\
                                             & SSAH                                & 4.44\%                & 24.44\%               \\
                                             & AdvSwap(ours)                       & \textbf{0.01}\%                & \textbf{0.28}\%                \\ \hline
\multirow{3}{*}{JPEG-30}                     & AdvDrop                             & \textbf{60.82}\%               & \textbf{59.66}\%               \\
                                             & SSAH                                & 83.02\%               & 88.98\%               \\
                                             & AdvSwap(ours)                       & 78.64\%               & 89.65\%               \\ \hline
\multirow{3}{*}{\begin{tabular}[c]{@{}c@{}}Shield\\ {[}20, 40, 60, 80{]}\end{tabular}} & AdvDrop                             & \textbf{54.57}\%               & \textbf{63.83}\%               \\
                                             & SSAH                                & 92.79\%               & 95.65\%               \\
                                             & AdvSwap(ours)                       & 89.32\%               & 95.90\%               \\ \hline
\end{tabular}}
\label{tab3}
\vspace{-6 mm}
\end{table}

Tab.~\ref{tab3} denotes that with no defense, AdvSwap achieves notably high attack success rates while maintaining stealthiness, as indicated by the low detection rates of 0.01\% and 0.28\% on the GTSRB and nuScenes datasets, respectively. This highlights the effectiveness of AdvSwap in generating adversarial perturbations that evade detection by the recognition system. 

Furthermore, when confronted with defense of JPEG-30 and Shield, AdvSwap demonstrates resilience by maintaining competitive attack success rates. For instance, under the JPEG-30 defense, AdvSwap achieves attack success rates of 78.64\% and 89.65\% on the GTSRB and nuScenes datasets, respectively. Although it is better than SSAH in terms of robustness, it still has a large gap compared with AdvDrop. It is due to the fact that too much emphasis is placed on attack concealment, the lower regularization parameter of the adversarial loss ($\lambda_{\text{adv}}$) sacrifices the robust performance of the algorithm.

In the second experiment, we investigate the impact of varying the $\lambda_{\text{adv}}$ on attack performance and stealthiness. Introducing the Fréchet Inception Distance (FID)~\cite{heusel2017gans} as a evaluation metric, we find that AdvSwap exhibits strong attack robustness with an FID value of 14. By adjusting $\lambda_{\text{adv}}$ to optimize attack performance while maintaining stealthiness, AdvSwap demonstrates its adaptability and effectiveness across different defense scenarios.

\begin{figure}[h!]
\centering
\includegraphics[width= 8.6 cm]{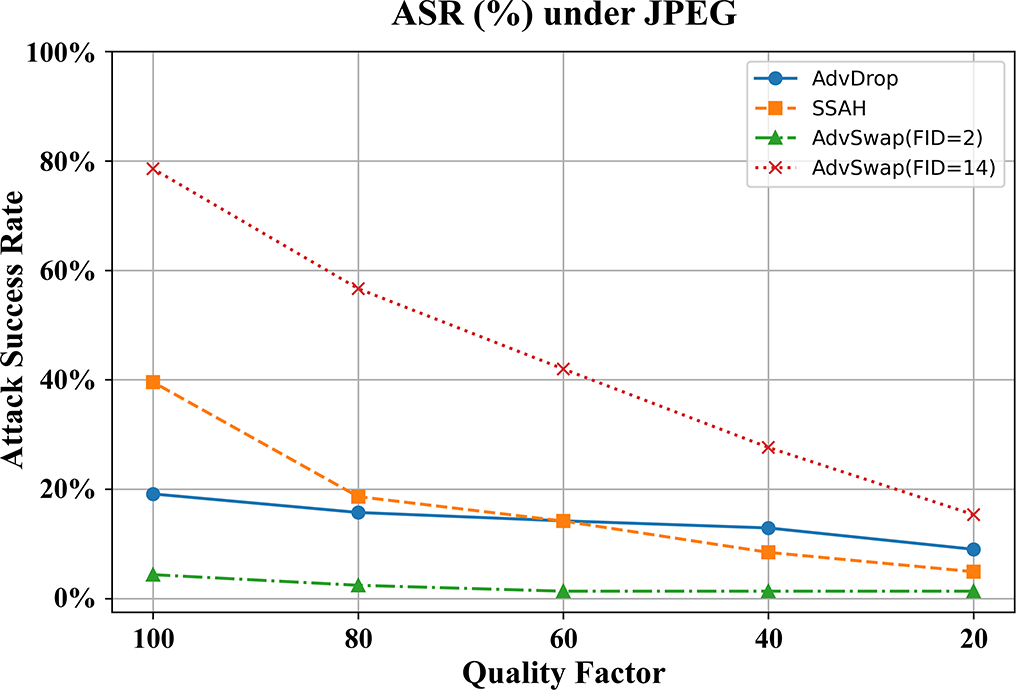}
\caption{JPEG defend with different quality factor.}
\label{6_roobust}
\vspace{-6 mm} 
\end{figure}

\subsection{Transferability Experiments}

We conducted a transferability experiments to assess the effectiveness of AdvSwap, our proposed adversarial attack method, on autonomous driving perception systems. Adversarial examples were generated using AdvSwap and evaluated on 3 different deep neural network architectures: VGG-16, ResNet-50, and Inception\_v3.

\begin{table}[htbp!]\small
\centering
\renewcommand\arraystretch{1.2}
\caption{Transferability Experiments}
\setlength{\tabcolsep}{3mm}{
\begin{tabular}{ccccc}
\hline
Classifier    & $L_{2}$ & LPIPS & SSIM  & ASR(\%) \\ 
\hline
VGG-16        & 0.687    & 0.012 & 0.997 & 100.00  \\
ResNet50      & 4.265    & 0.019 & 0.998 & 98.13   \\
Inception\_v3 & 3.648    & 0.015 & 0.997 & 99.26   \\ 
\hline
\end{tabular}}
\label{tab4}
\vspace{-6 mm}
\end{table}


Based on the results presented in Table \ref{tab4}, our attack algorithm, AdvSwap, demonstrates notable transferability across different classifiers. While exhibiting variations in \(L_2\), indicative of subtle sensitivities to different architectural designs, AdvSwap consistently achieved exceedingly high attack success rates (ASR ranging from 98.13\% to 100\%), thereby affirming its cross-model generality. Of particular note, the algorithm maintained near-perfect visual quality, as evidenced by SSIM values approaching unity and exceedingly low LPIPS scores(0.012, 0.019, and 0.015). We observe the Transferability and effectiveness of our method, as it can be seamlessly applied to various classifiers.

\subsection{Visualization and Analysis}

\begin{figure}[h!]
\centering
\includegraphics[width= 6 cm]{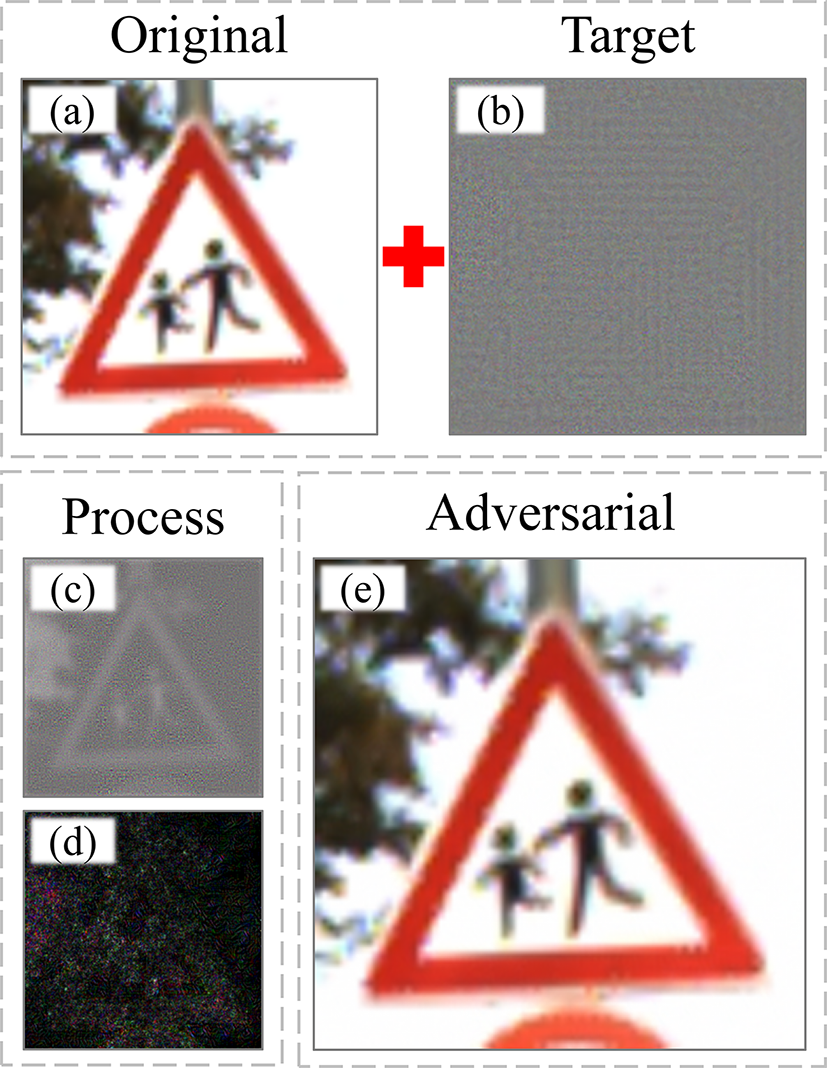}
\caption{Visualization of adversarial attack process.}
\label{07_visulization}
\vspace{-4 mm} 
\end{figure}

In Fig.\ref{07_visulization}, we depict the visualization of adversarial attack process. Firstly, in \ref{07_visulization}(b), we show the target image with reduced information generated by the adversarial optimizer. Subsequently, in \ref{07_visulization}(c), we observe a localized and refined exchange of information, particularly focusing on high-frequency components, preserving the overall structure and semantic content of the image. Furthermore, \ref{07_visulization}(d) displays the pixel differences between the adversarial sample and the original image, demonstrating a significant reduction compared to adversarial samples perturbed with direct noise injection. Finally, the generated adversarial sample in \ref{07_visulization}(e) exhibits enhanced stealthiness while maintaining adversarial efficacy, making it challenging for both humans and computers to detect the attack.

\section{CONCLUSIONS}

In this paper, we propose a new adversarial attack method based on high-frequency information exchange, named AdvSwap. This method extracts the high-frequency information of the image through wavelet transform, extracts the characteristics of each wavelet component through the residual network and inputs it into the reversible module. The reversible module built based on the reversible neural network can realize the same amount of high-frequency information exchange to completely retain and reply to the covert attack information. In addition, the Adversarial Optimizer and Classification Optimizer proposed in this article also bring high-quality guidance noise and optimization speed to this method. We conducted in-depth research and analysis on the proposed algorithm on two mainstream traffic data sets. Extensive experimental results show that the algorithm proposed in this paper can produce more covert adversarial samples compared with state-of-the-art algorithms. At the same time, the training parameters can be challenged according to needs to achieve good robustness and algorithm migration performance. 

\addtolength{\textheight}{-12cm}   



\section*{ACKNOWLEDGMENT}

The work is supported by Beijing Municipal Natural Science Foundation (No. L211003)



\end{document}